\documentclass[sigconf]{acmart}

\usepackage{amsmath}
\usepackage{booktabs}
\usepackage[linesnumbered,ruled,noend]{algorithm2e} 
\usepackage{multirow}
\usepackage{url}
\usepackage{bm}
\usepackage{amsmath,amsfonts}
\usepackage{textcomp}
\usepackage{xcolor}
\usepackage{amsthm}
\usepackage{subfigure}
\usepackage{stfloats}
\usepackage{caption}
\usepackage{wrapfig}

\SetKwInOut{Input}{Input}
\SetKwInOut{Output}{Output\,}
\SetKwInOut{Data}{Data}
\SetKwProg{Tree}{Tree}{}{EndTree}
\allowdisplaybreaks

\AtBeginDocument{%
  \providecommand\BibTeX{{%
    \normalfont B\kern-0.5em{\scshape i\kern-0.25em b}\kern-0.8em\TeX}}}

\setcopyright{iw3c2w3}
\begin{document}

\copyrightyear{2021}
\acmYear{2021}
\acmConference[WWW '21]{Proceedings of the Web Conference 2021}{April 19--23, 2021}{Ljubljana, Slovenia}
\acmBooktitle{Proceedings of the Web Conference 2021 (WWW '21), April 19--23, 2021, Ljubljana, Slovenia}
\acmPrice{}
\acmDOI{10.1145/3442381.3449999}
\acmISBN{978-1-4503-8312-7/21/04}

\title{OCT-GAN: Neural ODE-based Conditional Tabular GANs}

\author{Jayoung Kim, Jinsung Jeon, Jaehoon Lee, Jihyeon Hyeong, Noseong Park}
\email{{jayoung.kim, jjsjjs0902, ljh5694, jiji.hyeong, noseong}@yonsei.ac.kr}
\affiliation{%
  \institution{Yonsei University}
  \city{Seoul}
  \country{South Korea}
}

\renewcommand{\shortauthors}{Kim, et al.}

\begin{abstract}
Synthesizing tabular data is attracting much attention these days for various purposes. With sophisticate synthetic data, for instance, one can augment its training data. For the past couple of years, tabular data synthesis techniques have been greatly improved. Recent work made progress to address many problems in synthesizing tabular data, such as the imbalanced distribution and multimodality problems. However, the data utility of state-of-the-art methods is not satisfactory yet. In this work, we significantly improve the utility by designing our generator and discriminator based on neural ordinary differential equations (NODEs). After showing that NODEs have theoretically preferred characteristics for generating tabular data, we introduce our designs. The NODE-based discriminator performs a hidden vector evolution trajectory-based classification rather than classifying with a hidden vector at the last layer only. Our generator also adopts an ODE layer at the very beginning of its architecture to transform its initial input vector (i.e., the concatenation of a noisy vector and a condition vector in our case) onto another latent vector space suitable for the generation process. We conduct experiments with 13 datasets, including but not limited to insurance fraud detection, online news article prediction, and so on, and our presented method outperforms other state-of-the-art tabular data synthesis methods in many cases of our classification, regression, and clustering experiments.
\end{abstract}





\maketitle

\section{Introduction}
Many web-based applications use tabular data and many enterprise systems use relational database management systems~\cite{dbms}. For these reasons, many web-oriented researchers focus on various tasks on tabular data~\cite{10.1145/3308558.3313414,10.1145/3308558.3313629,10.1145/3308558.3313399,10.1145/3178876.3186029,10.1145/3366423.3380120,10.1145/3366423.3380174,10.1145/3366423.3380205,10.1145/3366423.3380181,10.1145/3366423.3380087,10.1145/3366423.3380039}. In this work, generating realistic synthetic tabular data is of our utmost interest. If the utility of synthetic data is reasonably high while being different enough from real data, it can greatly benefit many applications by enabling to use the synthetic data as (additional) training data~\cite{bowles2018gan,8622547,Choi2019SelfEnsemblingWG,antoniou2018data,chen2019faketables,tanaka2019data,10.1007/978-3-030-46133-1_23,tran2020data}. In one of our experiments, for instance, we synthesize a feature table extracted from raw online news articles to predict the number of shares (e.g., tweets and retweets) in social networks.

\begin{figure}
    \centering
    \subfigure[Overall Workflow]{\includegraphics[width=1.0\columnwidth]{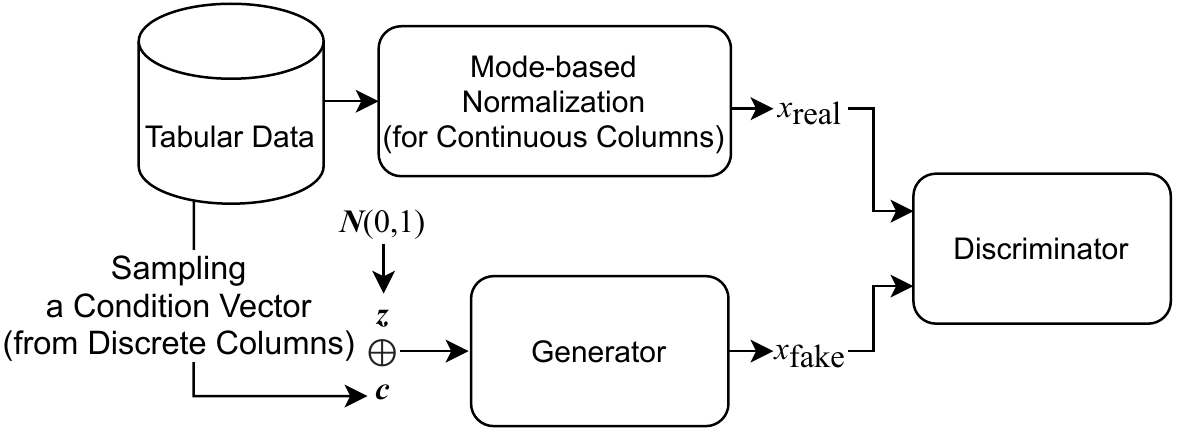}}
    \subfigure[Discriminator]{\includegraphics[width=1.0\columnwidth]{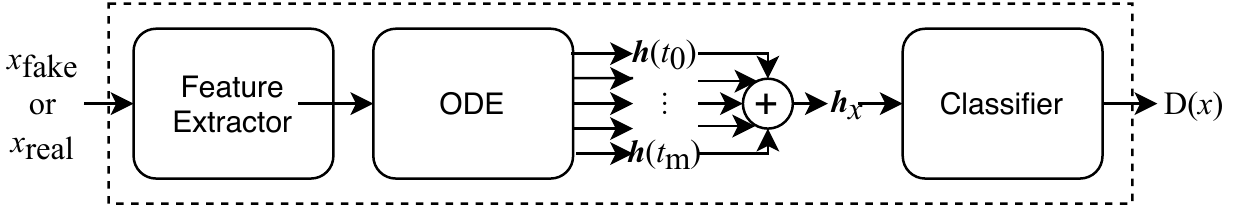}}
    \subfigure[Conditional Generator]{\includegraphics[width=0.7\columnwidth]{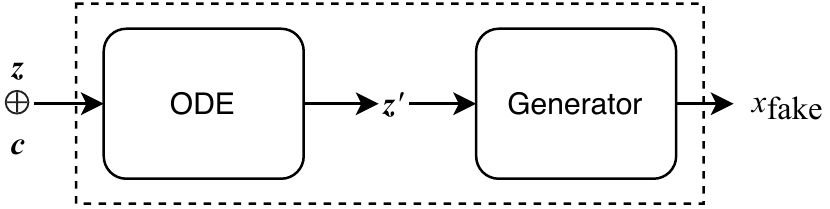}}
    \caption{Our proposed \texttt{OCT-GAN} architecture. (a) We preprocess raw tabular data with a mode-based normalization technique (See Section~\ref{sec:pre}). (b) The discriminator adopts an ODE layer to perform a trajectory-based classification (See Section~\ref{sec:disc}). (c) The ODE layer in the generator transforms $\bm{z} \oplus \bm{c}$, the concatenation of a noisy vector $\bm{z}$ and a condition vector $\bm{c}$, into another latent vector $\bm{z}'$ that will be fed into the generator (See Section~\ref{sec:gen}).}
    \label{fig:odegan}
\end{figure}

Generative adversarial networks (GANs), which consist of a generator and a discriminator, are one of the most successful generative models~\cite{NIPS2014_5423,NIPS2016_6125,pmlr-v70-arjovsky17a,10.5555/3295222.3295327,NIPS2018_7909}. GANs have been extended to various domains, ranging from images and texts to tables. Recently, Xu et al. introduced a tabular GAN, called \texttt{TGAN}, to synthesize tabular data~\cite{NIPS2019_8953}. \texttt{TGAN} now shows the state-of-the-art performance among existing GANs in generating tables in terms of \emph{model compatibility}, i.e., a machine learning model trained with synthetic (generated) data shows reasonable accuracy for unknown real test cases. In some cases, however, it is outperformed by non-GAN-based methods.


As noted in~\cite{DBLP:journals/corr/ChoiBMDSS17,DBLP:journals/corr/abs-1709-01648,DBLP:journals/corr/abs-1806-03384,NIPS2019_8953}, tabular data has irregular distribution and multimodality in many cases and existing techniques do not work. Based on the recent seminal work introducing Neural ODEs (NODEs)~\cite{zhuang2020adaptive,finlay2020train,2020arXiv200305271D,Quaglino2020SNODE}, to this end, we design a novel NODE-based conditional tabular GAN, called \texttt{OCT-GAN}. Fig.~\ref{fig:odegan} shows its detail design. In NODEs, a neural network $f$ learns a system of ordinary differential equations to approximate $\frac{d\bm{h}(t)}{dt}$, where $\bm{h}(t)$ is a hidden vector at time (or layer) $t$. Given a sample $x$ (i.e., a row or record in a table in our context), therefore, we can easily extract its hidden vector evolution trajectory from $t_0$ to $t_m$ while solving an integral problem, i.e., $\bm{h}(t_m) = \bm{h}(t_0) + \int_{t_0}^{t_m} f(\bm{h}(t),t;\bm{\theta}_f) dt$, where $\bm{\theta}_f$ means a set of parameters to learn for $f$. NODEs convert the integral problem into multiple steps of additions and we can retrieve a trajectory from those steps, i.e., $\{\bm{h}(t_0),\bm{h}(t_1),\bm{h}(t_2), \cdots, \bm{h}(t_m)\}$. Our discriminator equipped with a learnable ODE utilizes the extracted evolution trajectory to distinguish between real and synthetic samples (whereas other neural networks use only the last hidden vector, e.g., $\bm{h}(t_m)$ in the above example). This trajectory-based classification brings non-trivial freedom to the discriminator, making it be able to provide better feedback to the generator. Additional key part in our design is how to decide those time points $t_i$, for all $i$, to extract trajectories. We let the model learn them from data.

NODEs have one more characteristic in favor of our generator as well. That is, NODEs can be seen as a mapping function from $t_0$ to $t_m$ and the mapping function is always homeomorphic, i.e., bijective and continuous. We use this homeomorphic mapping to transform the initial input of our conditional generator, i.e., a noisy vector concatenated with a condition vector, into a latent vector in another vector space suitable for remaining procedures in the generator. A homeomorphic mapping does not drastically change input and the topology of input space is maintained in its output space. Therefore, we can maintain the original semantics of the noisy and condition vectors while mapping to another latent vector. One additional advantage of adopting a homeomorphic mapping is that it enables us to achieve smooth interpolations of generated fake samples.

We conduct experiments with 13 datasets for various machine learning tasks. Among all tested baseline methods to synthesize tabular data, the proposed \texttt{OCT-GAN} shows the best performance for many cases of our classification, regression, and clustering experiments. In addition, our method shows smooth interpolations of noisy vectors. Our contributions can be summarized as follows:
\begin{enumerate}
    \item Our discriminator has an ODE layer to extract a hidden vector evolution trajectory for classification.
    \item The trajectory is represented by a series of hidden vectors extracted at various layers (or time) $t_i$. We also train these extraction time points.
    \item The trajectory-based classification brings non-trivial benefits to the discriminator since we can use not only the last hidden vector but also all the information contained in the trajectory.
    \item Our generator adopts an initial ODE layer to transform $\bm{z} \oplus \bm{c}$ to another latent hidden vector $\bm{z}'$ suitable for the generation process (while maintaining the semantics contained in $\bm{z} \oplus \bm{c}$, i.e., a homeomorphic mapping).
    \item We conduct in-depth experiments with 13 datasets in total, ranging from insurance fraud detection to online news article spread prediction and so on. Our evaluation tasks include generating fake tabular data for likelihood estimation, classification, regression, and clustering, and our method outperforms existing methods by large margins in many cases.
\end{enumerate}

\section{Related Work}
We review the literature related to our work. We first introduce recent progress on GANs and then various tabular data synthesis techniques. We also describe NODEs in detail.

\subsection{Generative Adversarial Networks}
GANs consist of two neural networks: a generator and a discriminator. They perform a two-play zero-sum game and its equilibrium state is theoretically well defined, where the generator achieves the optimal generation quality and the discriminator cannot distinguish between real and fake samples. WGAN and its variants are widely used among many GANs proposed so far~\cite{pmlr-v70-arjovsky17a,10.5555/3295222.3295327,NIPS2018_7909}. In particular, WGAN-GP is one of the most successful models and is defined as follows:
\begin{align}\begin{split}\label{eq:wgan}
    \min_{\texttt{G}} \max_{\texttt{D}}\ & \mathbb{E}\big[\texttt{D}(\bm{x})\big]_{\bm{x} \sim p_{\bm{x}}}  -  \mathbb{E}\big[\texttt{D}(\texttt{G}(\bm{z}))\big]_{\bm{z} \sim p_{\bm{z}}}\\&- \lambda \mathbb{E} \big[(\| \nabla_{\bar{\bm{x}}} \texttt{D}(\bar{\bm{x}})\|_2 - 1)^2 \big]_{\bar{\bm{x}} \sim p_{\bar{\bm{x}}}},
\end{split}\end{align}where $p_{\bm{z}}$ is a prior distribution; $p_{\bm{x}}$ is a distribution of data; $\texttt{G}$ is a generator function; $\texttt{D}$ is a discriminator (or Wasserstein critic) function; $\bar{\bm{x}}$ is a randomly weighted combination of $\texttt{G}(\bm{z})$ and $\bm{x}$. The discriminator provides feedback on the quality of the generation. In addition, we let $p_g$ be a distribution of fake data induced by the function $\texttt{G}(\bm{z})$ from $p_{\bm{z}}$, and $p_{\bar{\bm{x}}}$ be a distribution created after the random combination. We typically use $\mathcal{N}(\bm{0},\bm{1})$ for the prior $p_{\bm{z}}$. Many task-specific GAN models are designed on top of the WGAN-GP framework. We use $\mathcal{L}_{\texttt{D}},\mathcal{L}_{\texttt{G}}$ to denote the WGAN-GP's loss functions to train the discriminator and the generator, respectively.

A popular variant of GANs are conditional GANs~\cite{mirza2014conditional,8100115}. Under the regime of conditional GANs, the generator $\texttt{G}(\bm{z}, \bm{c})$ is fed with a noisy vector $\bm{z}$ and a condition vector $\bm{c}$. In many cases, the condition vector is a one-hot vector denoting a class label to generate.

\subsection{Tabular Data Synthesis}
Tabular data synthesis, which generates a realistic synthetic table by modeling a joint probability distribution of columns in a table, encompasses many different methods depending on the types of data. For instance, Bayesian networks~\cite{avino2018generating, 10.1145/3134428} and decision trees~\cite{article} are used to generate discrete variables. A recursive modeling of tables using the Gaussian copula is used to generate continuous variables~\cite{7796926}. A differentially private algorithm for decomposition is used to synthesize spatial data~\cite{DBLP:journals/corr/abs-1103-5170, DBLP:journals/corr/ZhangXX16}. However, some constraints that these models have such as the type of distributions and computational problems have hampered high-fidelity data synthesis. 

In recent years, several data generation methods based on GANs have been introduced to synthesize tabular data, which mostly handle healthcare records. RGAN~\cite{esteban2017realvalued} generates continuous time-series healthcare records while \texttt{MedGAN}~\cite{DBLP:journals/corr/ChoiBMDSS17}, corrGAN~\cite{DBLP:journals/corr/abs-1804-00925} generate  discrete records. EhrGAN~\cite{DBLP:journals/corr/abs-1709-01648} generates plausible labeled records using semi-supervised learning to augment limited training data. \texttt{PATE-GAN}~\cite{Jordon2019PATEGANGS} generates synthetic data without endangering the privacy of original data. \texttt{TableGAN}~\cite{DBLP:journals/corr/abs-1806-03384}  improved tabular data synthesis using convolutional neural networks to maximize the prediction accuracy on the label column.

\subsection{Neural Ordinary Differential Equations}
Let $\bm{h}(t)$ be a function that outputs a hidden vector at time (or layer) $t$ in a neural network. In Neural ODEs (NODEs), a neural network $f$ with a set of parameters, denoted $\bm{\theta}_f$, approximates $\frac{d \bm{h}(t)}{d t}$, and $\bm{h}(t_m)$ is calculated by $\bm{h}(t_0)+\int_{t_0}^{t_m} f(\bm{h}(t), t;\bm{\theta}_f)\,dt$, where $f(\bm{h}(t), t;\bm{\theta}_f) = \frac{d \bm{h}(t)}{d t}$. In other words, the internal dynamics of the hidden vector evolution process is described by a system of ODEs parameterized by $\bm{\theta}_f$. One advantage of using NODEs is that we can interpret $t$ as continuous, which is discrete in usual neural networks. Therefore, more flexible constructions are possible in NODEs, which is one of the main reasons why we adopt an ODE layer in our discriminator.

To solve the integral problem, $\bm{h}(t_0)+\int_{t_0}^{t_m} f(\bm{h}(t), t;\bm{\theta}_f)\,dt$, in NODEs, we rely on an ODE solver which transforms an integral into a series of additions. The Dormand--Prince (DOPRI) method~\cite{DORMAND198019} is one of the most powerful integrators and is widely used in NODEs. It is a member of the Runge--Kutta family of ODE solvers. DOPRI dynamically controls its step size while solving an integral problem. It is now the default method in MATLAB, GNU Octave, etc.

\begin{figure}
    \centering
    \includegraphics[width=0.45\columnwidth]{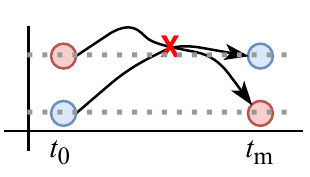}
    \caption{The locations of the red and blue points are swapped by the mapping from $t_0$ to $t_m$. NODEs cannot simultaneously learn the red and blue trajectories that cross each other since their topology (i.e., their relative positions) cannot be changed after a mapping in NODEs.}
    \label{fig:cross}
\end{figure}

Let $\phi_t : \mathbb{R}^{\dim(\bm{h}(t_0))} \rightarrow \mathbb{R}^{\dim(\bm{h}(t_m))}$ be a mapping from $t_0$ to $t_m$ created by an ODE after solving the integral problem. It is well-known that $\phi_t$ becomes a homeomorphic mapping: $\phi_t$ is continuous and bijective and $\phi_t^{-1}$ is also continuous for all $t \in [0,T]$, where $T$ is the last time point of the time domain~\cite{NIPS2019_8577,massaroli2020dissecting}. From this characteristic, the following proposition can be derived:
\begin{proposition}
The topology of the input space of $\phi_t$ is preserved in its output space, and therefore, trajectories crossing each other cannot be represented by NODEs, e.g., Fig.~\ref{fig:cross}.
\end{proposition}

While preserving the topology, NODEs can perform machine learning tasks and it was shown in~\cite{yan2020robustness} that it increases the robustness of representation learning to adversarial attacks.


Instead of the backpropagation method, the adjoint sensitivity method is used to train NODEs for its efficiency and theoretical correctness~\cite{NIPS2018_7892}. After letting $\bm{a}_{\bm{h}}(t) = \frac{d \mathcal{L}}{d \bm{h}(t)}$ for a task-specific loss $\mathcal{L}$, it calculates the gradient of loss w.r.t model parameters with another reverse-mode integral as follows:\begin{align*}\nabla_{\bm{\theta}_f} \mathcal{L} = \frac{d \mathcal{L}}{d \bm{\theta}_f} = -\int_{t_m}^{t_0} \bm{a}_{\bm{h}}(t)^{\mathtt{T}} \frac{\partial f(\bm{h}(t), t;\bm{\theta}_f)}{\partial \bm{\theta}_f} dt.\end{align*}

$\nabla_{\bm{h}(0)} \mathcal{L}$ can also be calcualte in a similar way and we can propagate the gradient backward to layers earlier than the ODE if any. It is worth of mentioning that the space complexity of the adjoint sensitivity method is $\mathcal{O}(1)$ whereas using the backpropagation to train NODEs has a space complexity proportional to the number of DOPRI steps. Their time complexities are similar or the adjoint sensitivity method is slightly more efficient than that of the backpropagation. Therefore, we can train NODEs efficiently.




\section{Proposed Methods}
In this section, we describe our \texttt{OCT-GAN}. We first describe our data preprocessing method, and then describe both our discriminator and generator. We adopt an ODE layer for the following reasons in each of the discriminator and the generator:
\begin{enumerate}
    \item In the discriminator, we can interpret time (or layer) $t$ as continuous in its ODE layer. We can also perform trajectory-based classification by finding optimal time points that lead to improved classification performance.
    \item In the conditional generator, we exploit the homeomorphic characteristic of NODEs to transform $\bm{z}\oplus\bm{c}$ onto another latent space while preserving the (semantic) topology of the initial latent space. We propose to use this because i) a data distribution in tabular data is irregular and difficult to directly capture~\cite{NIPS2019_8953} and ii) by finding an appropriate latent space the generator can generate better samples~\cite{Karras_2019_CVPR}. At the same time, interpolating noisy vectors given a fixed condition can be smooth.
    \item Therefore, the entire generation process can be separated into the following two stages as in Fig.~\ref{fig:trans}: 1) transforming the initial input space into another latent space (potentially close to a real data distribution) while maintaining the topology of the input space, and 2) the remaining generation process finds a fake distribution matched to the real data distribution.
\end{enumerate}

\subsection{Preprocessing of Tabular Data}\label{sec:pre}
We consider tabular data with two types of columns: discrete columns, denoted \{$D_1$, $D_2$, ..., $D_{N_D}$\}, and continuous columns, denoted \{$C_1$, $C_2$, ..., $C_{N_C}$\}. Discrete values are transformed to one-hot vectors as usual. However, continuous values are preprocessed with a mode-specific normalization technique~\cite{NIPS2019_8953}. GANs generating tabular data frequently suffer from mode collapse and irregular data distribution. By specifying modes before training, the mode-specific normalization can alleviate the problems. The $i_{th}$ raw sample $r_i$ (a row or record in the tabular data) can be written as $d_{i,1}$ $\oplus$ $d_{i,2}$ $\oplus$ ... $\oplus$ $d_{i,N_D}$ $\oplus$ $c_{i,1}$ $\oplus$ $c_{i,2}$ $\oplus$ ... $\oplus$ $c_{i,N_C}$, where $d_{i,j}$ (resp. $c_{i,j}$) is a value in column $D_j$ (resp. column $C_j$). After the following three steps, the raw sample $r_i$ is preprocessed to $x_i$. 
\begin{enumerate}
    \item Each discrete values \{$d_{i,1}$, $d_{i,2}$, ..., $d_{i,N_D}$\} are transformed to one-hot vector \{$\bm{d}_{i,1}$, $\bm{d}_{i,2}$, ..., $\bm{d}_{i,N_D}$\}.

    \item Using the variational Gaussian mixture (VGM) model, we fit each continuous column $C_j$ to a Gaussian mixture. The fitted Gaussian mixture is $\Pr_{j}(c_{i,j})= \sum_{k=1}^{n_j}w_{j,k}\mathcal{N}(c_{i,j};\mu_{j,k}, \sigma_{j,k})$, where $n_j$ is the number of modes (i.e., the number of Gaussian distributions) in columns $C_j$. $w_{j,k}, \mu_{j,k},\sigma_{j,k}$ are a fitted weight, mean and standard deviation of $k_{th}$ Gaussian distribution.

    \item With a probability of $\Pr_{j}(k)= \frac{w_{j,k}\mathcal{N}(c_{i,j};\mu_{j,k}, \sigma_{j,k})}{\sum_{p=1}^{n_j}w_{j,p}\mathcal{N}(c_{i,j};\mu_{j,p}, \sigma_{j,p})}$, we sample an appropriate mode $k$ for $c_{i,j}$. We then normalize $c_{i,j}$ from the mode $k$ with its fitted standard deviation, and save the normalized value $\alpha_{i,j}$ and the mode information $\beta_{i,j}$. For example, if there are 4 modes and we pick the third mode, i.e., $k=3$, then $\alpha_{i,j}$ is $\frac{c_{i,j} - \mu_3}{4\sigma_3}$ and $\beta_{i,j}$ is $[0,0,1,0]$.

 	\item As a result, $r_i$ is transformed to $x_i$ which is denoted as follows:$$x_i = \alpha_{i,1} \oplus \beta_{i,1} \oplus \cdots \oplus \alpha_{i,N_c} \oplus \beta_{i,N_c} \oplus \bm{d}_{i,1} \oplus \cdots \oplus\bm{d}_{i,N_D}.$$
\end{enumerate}
We note that in $x_i$, we can specify the detailed mode-based information of $r_i$. Our discriminator and generator work with $x_i$ instead of $r_i$ for its clarification on modes. However, $x_i$ can be readily changed to $r_i$, once generated, using the fitted parameters of the Gaussian mixture.

\subsection{Discriminator}\label{sec:disc}
We design a NODE-based discriminator and consider the trajectory of $\bm{h}(t)$, where $t \in [0,t_m]$, when predicting whether an input sample $x$ is real or fake. To this end, we use the following ODE-based discriminator that outputs $\texttt{D}(x)$ given a (preprocessed or generated) sample $x$:
\begin{align}
    \bm{h}(0) &= \mathtt{Drop}(\mathtt{Leaky}(\mathtt{FC2}(\mathtt{Drop}(\mathtt{Leaky}(\mathtt{FC1}(x)))))),\\
    \bm{h}(t_1) &= \bm{h}(0)+\int_{0}^{t_1} f(\bm{h}(0),t;\bm{\theta}_f)\,dt,\label{eq:ode1}\\
    \bm{h}(t_2) &= \bm{h}(t_1)+\int_{t_1}^{t_2} f(\bm{h}(t_1),t;\bm{\theta}_f)\,dt,\\
    &\vdots\\
    \bm{h}(t_m) &= \bm{h}(t_{m-1})+\int_{t_{m-1}}^{t_m} f(\bm{h}(t_{m-1}),t;\bm{\theta}_f)\,dt,\label{eq:ode2}\\
    \bm{h}_x &= \bm{h}(0) \oplus \bm{h}(t_1) \oplus \bm{h}(t_2) \oplus \cdots \oplus \bm{h}(t_m),\\
    \texttt{D}(x) &= \mathtt{FC5}(\mathtt{Leaky}(\mathtt{FC4}(\mathtt{Leaky}(\mathtt{FC3}(\bm{h}_x))))),\label{eq:d:cla}
\end{align}where $\oplus$ means the concatenation operator, $\mathtt{Leaky}$ is the leaky ReLU, $\mathtt{Drop}$ is the dropout, and $\mathtt{FC}$ is the fully connected layer. The ODE function $f(\bm{h}(t),t;\bm{\theta}_f)$ is defined as follows:
\begin{align}
    \mathtt{ReLU}(\mathtt{BN}(\mathtt{FC7}(\mathtt{ReLU}(\mathtt{BN}(\mathtt{FC6}(\mathtt{ReLU}(\mathtt{BN}(\bm{h}(t)))\oplus t)))))),
\end{align}where $\mathtt{BN}$ is the batch normalization and $\mathtt{ReLU}$ is the rectified linear unit.

We note that the trajectory of $\bm{h}(t)$ is continuous in NODEs. However, it is difficult to consider continuous trajectories in training GANs. To discretize the trajectory of $\bm{h}(t)$, therefore, $t_1, t_2, \cdots, t_m$ are trained and $m$ is a hyperparameter in our proposed method. We also note that Eqs.~\eqref{eq:ode1} to~\eqref{eq:ode2} share the same parameter $\bm{\theta}_f$, which means they constitute a single system of ODEs but for the purpose of discretization we separate them. After letting $\bm{a}_t(t) = \frac{d \mathcal{L}}{d t}$, we use the following gradient definition (derived from the adjoint sensitivity method) to train $t_i$ for all $i$:
\begin{proposition}
The gradient of loss $\mathcal{L}$ w.r.t. $t_m$ can be calculated in the following way:
$$\nabla_{t_m} \mathcal{L} = \frac{d \mathcal{L}}{d t_m} = \bm{a}_{\bm{h}}(t_m) f(\bm{h}(t_m), t_m;\bm{\theta}_f).$$
\end{proposition}
\begin{proof}
First, because $\bm{a}_{\bm{h}}(t_m) = \frac{d \mathcal{L}}{d \bm{h}(t_m)}$ by its definition,$$\bm{a}_{\bm{h}}(t_m) f(\bm{h}(t_m), t_m;\bm{\theta}_f) = \frac{d \mathcal{L}}{d \bm{h}(t_m)} f(\bm{h}(t_m), t_m;\bm{\theta}_f),$$and then by the definition of $f$, $$\frac{d \mathcal{L}}{d \bm{h}(t_m)} f(\bm{h}(t_m), t_m;\bm{\theta}_f) = \frac{d \mathcal{L}}{d \bm{h}(t_m)} \frac{d \bm{h}(t_m)}{d t_m} = \frac{d \mathcal{L}}{d t_m}.$$
\end{proof}

For the same reason above, $\nabla_{t_i} \mathcal{L}=\frac{d \mathcal{L}}{d t_i} = \bm{a}_{\bm{h}}(t_i) f(\bm{h}(t_i), t_i;\bm{\theta}_f)$ where $i < m$. But we do not want to save any intermediate adjoint states for space complexity purposes and calculate the gradient with a reverse-mode integral as follows:
\begin{align*}
\nabla_{t_i} \mathcal{L} = \bm{a}_{\bm{h}}(t_m) f(\bm{h}(t_m), t_m;\bm{\theta}_f) -\int_{t_m}^{t_i} \bm{a}_{\bm{h}}(t) \frac{\partial f(\bm{h}(t), t;\bm{\theta}_f)}{\partial t} dt.
\end{align*}
We note that we need to save only one adjoint sate $\bm{a}_{\bm{h}}(t_m)$ and calculate $\nabla_{t_i} \mathcal{L}$ with the two functions $f$ and $\bm{a}_{\bm{h}}(t)$.

\begin{figure}
    \centering
    \includegraphics[width=0.45\columnwidth]{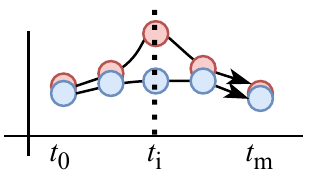}
    \caption{Suppose that the two red/blue trajectories from $t_0$ to $t_m$ are all similar except around $t_i$. Because we train such distinguishing time points, our trajectory-based classification can correctly classify them. Without training $t_i$, those intermediate time points should be set by user, which is sub-optimal.}
    \label{fig:ex}
\end{figure}
\begin{figure}
    \centering
    \includegraphics[width=0.45\columnwidth]{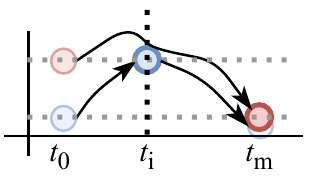}
    \caption{The red and blue trajectories do not cross each other and can be learned by NODEs. By taking the blue hidden vector at $t_i$ and the red hidden vector at $t_m$, however, we can swap their positions, which is impossible in Fig.~\ref{fig:cross}. Therefore, our trajectory-based classification is necessary to improve NODEs.}
    \label{fig:overcome}
\end{figure}

We typically use the last hidden vector $\bm{h}(t_m)$ for classification. In our case, however, we use the entire trajectory for classification. When using only the last hidden vector, all needed information for classification should be correctly captured in it. In our setting, however, even two similar last hidden vectors can be easily distinguished if their intermediate trajectories are different at least at a value of $t$. In addition, we train $t_i$, which further improves the efficacy of the proposed method by finding key time points to distinguish trajectories. We note that training $t_i$ is impossible in usual neural networks because their layer constructions are discrete. Fig.~\ref{fig:ex} illustrates such an example that only our NODE-based discriminator with learnable intermediate time points can correctly classify, and Fig.~\ref{fig:overcome} also shows that our method can address the problem of the limited learning representation of NODEs.

\subsection{Conditional Generator}\label{sec:gen}
The proposed \texttt{OCT-GAN} is a conditional GAN and its generator reads a noisy vector as well as a condition vector to generate a fake sample. Our definition of the condition vector is as follows:$$\bm{c} =\bm{c_1} \oplus \cdots \oplus\bm{c}_{N_D},$$where $\bm{c}_i$ is either a zero vector or a random one-hot vector of the $i_{th}$ discrete column. We randomly decide $s\in\{1,2,\cdots,N_D\}$ and only $\bm{c}_s$ is a random one-hot vector and for all other $i \neq s$, $\bm{c}_i$ is a zero vector, i.e., we specify a discrete value in the $s_{th}$ discrete column.

Given an initial input $\bm{p}(0) = \bm{z} \oplus \bm{c}$, we feed it into an ODE layer to transform into another latent vector. We denote this transformed vector by $\bm{z}'$. For this transformation, we use an ODE layer independent from the ODE layer in the discriminator as follows:$$\bm{z}' = \bm{p}(1) = \bm{p}(0) + \int_0^1 g(\bm{p}(t), t; \bm{\theta}_g) dt.$$

We fix the integral time to $[0,1]$ because any ODE in $[0,w]$, $w>0$, with $g$ can be reduced into a unit-time integral with $g'$ by letting $g' = \frac{g(\bm{p}(t), t; \bm{\theta}_g)}{w}$.

As noted earlier, an ODE is a homeomorphic mapping. We exploit the characteristic to design a semantically reliable mapping (or transformation). GANs typically use a noisy vector sampled from a Gaussian distribution, which is known as sub-optimal~\cite{Karras_2019_CVPR}. Thus, the transformation is needed in our case.

The Gr\"onwall–Bellman inequality states that given an ODE $\phi_t$ and its two initial states $\bm{p}_1(0) = \bm{x}$ and $\bm{p}_2(0) = \bm{x} + \bm{\delta}$, there exists a constant $\tau$ such that $\|\phi_t(\bm{x}) - \phi_t(\bm{x} + \bm{\delta})\| \leq exp(\tau)\|\bm{\delta}\|$~\cite{pachpatte1997inequalities}. In other words, two similar input vectors with small $\bm{\delta}$ will be mapped to close to each other within a boundary of $exp(\tau)\|\bm{\delta}\|$.

\begin{figure}[t]
    \centering
    \includegraphics[width=1\columnwidth]{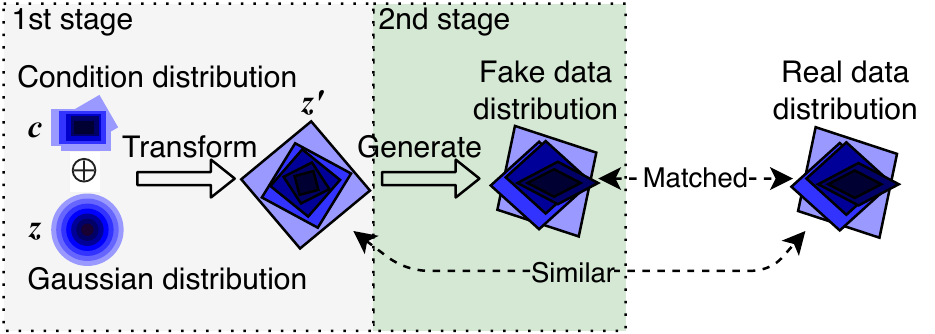}
    \caption{The ODE layer in the generator transforms the concatenation of a noisy vector and a condition vector, $\bm{z} \oplus \bm{c}$, into $\bm{z}'$. The 1st transformation is a homeomorphic mapping that maintains the original (semantic) topology whereas the 2nd generation process is not homeomorphic.}
    \label{fig:trans}
\end{figure}

In addition, we do not extract $\bm{z}'$ from intermediate time points so the generator's ODE learns a homeomorphic mapping. Thus, the topology of the initial input vector space is maintained. The initial input vector $\bm{p}(0)$ contains non-trivial information on what to generate, e.g., condition, so we would like to maintain the relationships among initial input vectors while transforming them onto another latent vector space suitable for generation. Fig.~\ref{fig:trans} shows an example of our two-stage approach where i) the ODE layer finds a balancing distribution between the initial input distribution and the real data distribution and ii) the following procedures generate realistic fake samples. In particular, our transformation makes the interpolation of synthetic samples smooth, i.e., given two similar initial inputs, two similar synthetic samples are generated by the generator (as proved in the Gr\"onwall–Bellman inequality) --- we show these smooth interpolations in our experiment section. The proposed generator equipped with learning the optimal transformation is as follows:
\begin{align}
    \bm{p}(0) &= \bm{z} \oplus \bm{c},\\
    \bm{z}' &= \bm{p}(0)+\int_{0}^{1} g(\bm{p}(t),t;\bm{\theta}_g)\,dt,\\
    \bm{h}(0) &= \bm{z}' \oplus \mathtt{ReLU}(\mathtt{BN}(\mathtt{FC1}(\bm{z}'))),\\
    \bm{h}(1) &= \bm{h}(0) \oplus \mathtt{ReLU}(\mathtt{BN}(\mathtt{FC2}(\bm{h}(0)))),\\
    \hat{\alpha}_i &= \mathtt{Tanh}(\mathtt{FC3}(\bm{h}(1))), 1 \leq i \leq N_c,\\
    \hat{\beta}_i &= \mathtt{Gumbel}(\mathtt{FC4}(\bm{h}(1))), 1 \leq i \leq N_c,\\
    \hat{\bm{d}}_j &= \mathtt{Gumbel}(\mathtt{FC5}(\bm{h}(1))), 1 \leq j \leq N_d,
\end{align}where $\mathtt{Tanh}$ is the hyperbolic tangent, and $\mathtt{Gumbel}$ is the Gumbel-softmax to generate one-hot vectors. The ODE function $g(\bm{p}(t),t;\bm{\theta}_g)$ is defined as follows:
\begin{align}
    \overset{\textrm{8 layers of $\mathtt{FC}$ and  $\mathtt{Leaky}$}}{\overbrace{\mathtt{Leaky}(\mathtt{FC13}(\cdots\mathtt{Leaky}(\mathtt{FC6}}}(\mathtt{Norm}(\bm{p}(t)) \oplus t))\cdots)),
\end{align}where $\mathtt{Norm}(\bm{p})=\frac{\bm{p}}{\|\bm{p}\|_2}$.



As stated earlier, we specify a discrete value in a discrete column as a condition. Thus, it is required that $\hat{\bm{d}}_s = \bm{c}_s$, and we use a cross-entropy loss to enforce the match, denoted $\mathcal{L}_{matching}=H(\bm{c}_s,\hat{\bm{d}}_s)$. Another possible design choice is to copy $\bm{c}_s$ to $\hat{\bm{d}}_s$. However, we do not copy for a principled training of the generator.

\subsection{Training Algorithm}
We train \texttt{OCT-GAN} using the loss in Eq.~\eqref{eq:wgan} in conjunction with $\mathcal{L}_{matching}$ and its training algorithm is in Alg.~\ref{alg1}. To train \texttt{OCT-GAN}, we need a real table $\mathtt{T}_{\textrm{train}}$, and a maximum epoch number $max\_epoch$. After creating a mini-batch $b$ (line~\ref{alg1:1}), we perform the adversarial training (lines~\ref{alg1:2},~\ref{alg1:3}), followed by updating $t_i$ with the custom gradient calculated by the adjoint sensitivity method (line~\ref{alg1:4}).

\begin{algorithm}[t]
\DontPrintSemicolon
  \caption{How to train \texttt{OCT-GAN}}\label{alg1}
  \Input{A training table $\mathtt{T}_{\textrm{train}}$; a max epoch $max\_epoch$; learning rates $\lambda_\texttt{G},\lambda_\texttt{D},\lambda_t$}
  \Output{A trained generator}
  Initialize a generator $\texttt{G}$ and a discriminator $\texttt{D}$\;
  $k \gets 0$\;
  \While{$ k < max\_epoch$}{
  \For{each mini-batch $b \in \mathtt{T}_{\textrm{train}}$}{\label{alg1:1}
    \tcc{Perform adversarial training.}
    Train the discriminator $\texttt{D}$ with mini-batch $b$, learning rate $\lambda_\texttt{D}$, and loss $\mathcal{L}_{\texttt{D}}$ \label{alg1:2}\;
    Train the generator $\texttt{G}$ with mini-batch $b$, learning rate $\lambda_G$, loss $\mathcal{L}_{\texttt{G}} + \mathcal{L}_{matching}$ \label{alg1:3}\;
    \tcc{Update the intermediate time points.}
    $t_i \gets t_i - \lambda_t \nabla_{t_i} \mathcal{L}$, for all $i$\label{alg1:4}\;
  }
  $k \gets k + 1$\;
  Anneal $\lambda_\texttt{G},\lambda_\texttt{D},\lambda_t$ with a decay factor of $\psi$ every $\xi$ epoch\;
  }
  \Return the trained generator $\texttt{G}$
\end{algorithm}

The space complexity to calculate $\nabla_{t_i} \mathcal{L}$ is $\mathcal{O}(1)$ (see Section~\ref{sec:disc}). Calculating $\nabla_{t_j} \mathcal{L}$ subsumes the computation of $\nabla_{t_i} \mathcal{L}$, where $t_0 \leq t_j < t_i \leq t_m$. While solving the reverse-mode integral from $t_m$ to $t_0$, thus, we can retrieve $\frac{d \mathcal{L}}{d t_i}$ for all $i$. Therefore, the space complexity to calculate all the gradients is $\mathcal{O}(m)$ at line~\ref{alg1:4}, which is additional overhead incurred by our method.

\begin{table*}[t]
\small
\caption{\#C, \#B, and \#M mean the number of continuous columns, binary columns and multi-class discrete columns, respectively. C and R in Task mean classification and regression, respectively.}\label{tbl:sta}
\begin{tabular}{cc|ccccccccc||c|ccccccccccc}
\specialrule{1pt}{1pt}{1pt}
& \multirow{2}{*}{Name}         & & \multicolumn{7}{c}{Simulated Data} & & \multirow{2}{*}{Name} & \multicolumn{10}{c}{Real Data} &\\ \cline{4-10} \cline{14-22}
& & & \multicolumn{1}{c}{\#train/test} & & \multicolumn{1}{c}{\#C} & & \multicolumn{1}{c}{\#B} & & \multicolumn{1}{c}{\#M} & & & & \multicolumn{1}{c}{\#train/test} & & \#C & & \#B & & \#M & & Task &\\ \specialrule{1pt}{1pt}{1pt}
& \texttt{Grid}     & & 10k/10k                         & & 2                      & & 0                      & & 0                      & & \texttt{Adult} & & 23k/10k                         & & 6  & & 2  & & 7  & & C   &\\
& \texttt{Gridr}    & & 10k/10k                         & & 2                      & & 0                      & & 0                      & & \texttt{Census}              & & 200k/100k                       & & 7  & & 3  & & 31 & & C   &\\
& \texttt{Ring}     & & 10k/10k                         & & 2                      & & 0                      & & 0                      & & \texttt{Covertype}           & & 481k/100k                       & & 10 & & 44 & & 1  & & C   &\\
& \texttt{Asia}     & & 10k/10k                         & & 0                      & & 8                      & & 0                      & & \texttt{Credit}              & & 264k/20k                        & & 29 & & 1  & & 0  & & C   &\\
& \texttt{Alarm}    & & 10k/10k                         & & 0                      & & 13                     & & 24                     & & \texttt{Intrusion}           & & 394k/100k                       & & 26 & & 5  & & 10 & & C   &\\
& \texttt{Child}    & & 10k/10k                         & & 0                      & & 8                      & & 12                     & & \texttt{News}                & & 31k/8k                          & & 45 & & 14 & & 0  & & R   &\\
& \texttt{Insurance} & & 10k/10k                         & & 0                      & & 8                      & & 19                     & &  & &         & &    & &    & &    & &   &  \\ \specialrule{1pt}{1pt}{1pt}
\end{tabular}
\end{table*}

\section{Experiments}
We describe our experimental environments and results for likelihood estimation, classification, regression, clustering, and so on.

\subsection{Likelihood Estimation with Simulated Data}
\subsubsection{Data}
We first conduct experiments with simulated datasets. We collected various pre-trained Bayesian networks and Gaussian mixture models from the literature. Using the pre-trained models (or oracles), we generate $\mathtt{T}_{\textrm{train}}$ and $\mathtt{T}_{\textrm{test}}$, each of which is used for training and testing, respectively: \texttt{Grid} and \texttt{Ring} are from the Gaussian mixture models~\cite{NIPS2017_6923} and \texttt{Alarm}, \texttt{Child}, \texttt{Asia}, and \texttt{Insurance} are from the Bayesian networks~\cite{bnr} as summarized in Table~\ref{tbl:sta}.

\subsubsection{Evaluation Methodology}
One advantage of using the simulated data is that we can estimate the \emph{likelihood fitness} of synthetic data given an oracle (pre-trained model) $\mathcal{S}$. The overall evaluation workflow is as follows:
\begin{enumerate}
    \item Using $\mathtt{T}_{\textrm{train}}$, we train generative models including \texttt{OCT-GAN}.
    \item We generate synthetic data from each trained generative model. Let $\mathcal{F}$ be this synthetic data.
    \item We measure the likelihood of $\mathcal{F}$ given $\mathcal{S}$, denoted $\Pr(\mathcal{F} | \mathcal{S})$.
    \item We train another oracle $\mathcal{S}'$ with $\mathcal{F}$ from scratch.
    \item We measure $\Pr(\mathtt{T}_{\textrm{test}} | \mathcal{S}')$, the likelihood of $\mathtt{T}_{\textrm{test}}$ given $\mathcal{S}'$.
\end{enumerate}

We note that the two likelihood estimates should be good enough at the same time. A low value for $\Pr(\mathtt{T}_{\textrm{test}} | \mathcal{S}')$ means that $\mathcal{F}$ contains limited cases of $\mathtt{T}_{\textrm{train}}$, i.e., mode collapse. We repeat the generation-testing experiments ten times to find their average performance, which is also the case in the remaining experiments in this paper.

\subsubsection{Baseline Methods}
We consider the following baselines: i) The case where we use $\mathtt{T}_{\textrm{train}}$ instead of $\mathcal{F}$ is shown in the row titled ``$\mathtt{T}_{\textrm{train}}$'' in Tables~\ref{tbl:gau} and~\ref{tbl:bn}; ii) \texttt{CLBN}~\cite{1054142} is a Bayesian network built by the Chow-Liu algorithm representing a joint probability distribution; iii) \texttt{PrivBN}~\cite{10.1145/3134428} is a differentially private method for synthesizing tabular data using Bayesian networks; iv) \texttt{MedGAN}~\cite{DBLP:journals/corr/ChoiBMDSS17} is a GAN that generates discrete medical records by incorporating non-adversarial losses; v) \texttt{VEEGAN}~\cite{NIPS2017_6923} is a GAN that generates tabular data with an additional reconstructor network to avoid mode collapse; vi) \texttt{TableGAN}~\cite{DBLP:journals/corr/abs-1806-03384} is a GAN that generates tabular data using convolutional neural networks; vii) \texttt{TVAE}~\cite{ishfaq2018tvae} is a variational autoencoder (VAE) model to generate tabular data; viii) \texttt{TGAN}~\cite{NIPS2019_8953} is a GAN that generates tabular data with mixed types of variables. We use these baselines' hyperparameters recommended in their original paper and github repositories.

\subsubsection{Hyperparameters}
We test the following sets of hyperparameters: $\lambda_t = \lambda_G = \lambda_D = \{2e-3, 2e-4, 2e-5, 2e-6\}$, the mini-batch size is \{500, 1000, 1500\}, and the number of intermediate time points to train in the discriminator is $m = \{3, 5, 7\}$. The maximum epoch number is $max\_epoch = 300$. The (input, output) dimensionality of each layer is as follows:
\begin{enumerate}
    \item In the discriminator,
    \begin{enumerate}
        \item $(\dim(x), 256)$ for $\mathtt{FC1}$,
        \item $(256,256)$ for $\mathtt{FC2}$,
        \item $(\dim(\bm{h}_x),2\dim(\bm{h}_x))$ for $\mathtt{FC3}$,
        \item $(2\dim(\bm{h}_x),\dim(\bm{h}_x))$ for $\mathtt{FC4}$,
        \item $(\dim(\bm{h}_x),1)$ for $\mathtt{FC5}$,
        \item $(\dim(\bm{h}(x))+1,\dim(\bm{h}(x)))$ for $\mathtt{FC6}$,
        \item $(\dim(\bm{h}(x)),\dim(\bm{h}(x)))$ for $\mathtt{FC7}$.
    \end{enumerate}
    \item In the generator,
    \begin{enumerate}
        \item $(\dim(\bm{p}(0)), 256)$ for $\mathtt{FC1}$,
        \item $(256,256)$ for $\mathtt{FC2}$,
        \item $(512,1)$ for $\mathtt{FC3}$,
        \item $(512,n_i)$ for $\mathtt{FC4}$,
        \item $(512,\dim(\bm{d}_i))$ for $\mathtt{FC5}$,
        \item $(\dim(\bm{p}(x))+1,\dim(\bm{p}(x)))$ for $\mathtt{FC6}$,
        \item $(\dim(\bm{p}(x)),\dim(\bm{p}(x)))$ for $\mathtt{FC7}$ to $\mathtt{FC13}$.
    \end{enumerate}
\end{enumerate}

The number of modes in VGM is $n_j = \{10, 20, 30\}$. The learning rate decay factor and period are $\psi=0.97$, $\xi = 2$ so there are almost no updates around 300 epochs and the model converges. All experiments were conducted in the following software and hardware environments: \textsc{Ubuntu} 18.04 LTS, \textsc{Python} 3.6.6, \textsc{Numpy} 1.18.5, \textsc{Scipy} 1.5, \textsc{Matplotlib} 3.3.1, \textsc{PyTorch} 1.2.0, \textsc{CUDA} 10.0, and \textsc{NVIDIA} Driver 417.22, i9 CPU, and \textsc{NVIDIA RTX Titan}.

\subsubsection{Experimental Results}
In Tables~\ref{tbl:gau} and~\ref{tbl:bn}, all likelihood estimation results are included. \texttt{CLBN} and \texttt{PrivBN} show fluctuating performance. \texttt{CLBN} and \texttt{PrivBN} are good in \texttt{Ring} and \texttt{Asia}, respectively while \texttt{PrivBN} shows poor performance in \texttt{Grid}, and \texttt{Gridr}. \texttt{TVAE} shows good performance for $\Pr(\mathcal{F} | \mathcal{S})$ in many cases but relatively worse performance than others for $\Pr(\mathtt{T}_{\textrm{test}} | \mathcal{S}')$ in \texttt{Grid} and \texttt{Insurance}, which means mode collapse. At the same time, \texttt{TVAE} shows nice performance for \texttt{Gridr}. All in all, \texttt{TVAE} shows reasonable performance in these experiments.

Among many GAN models except \texttt{OCT-GAN}, \texttt{TGAN} and \texttt{TableGAN} show reasonable performance, and other GANs are inferior to them in many cases, e.g., -14.3 for \texttt{TableGAN} vs. -14.8 for \texttt{TGAN} vs. -18.1 for \texttt{VEEGAN} in \texttt{Insurance} with $\Pr(\mathtt{T}_{\textrm{test}} | \mathcal{S}')$. However, all these models are significantly outperformed by our proposed \texttt{OCT-GAN}. In all cases, \texttt{OCT-GAN} is better than \texttt{TGAN}, the state-of-the-art GAN model.

\begin{table*}[t]
\small
\centering
\setlength{\tabcolsep}{3pt}
\caption{Likelihood estimation with Gaussian mixture models. The best (resp. the second best) results are highlighted in boldface (resp. with underline). Our \texttt{OCT-GAN} outperforms \texttt{TGAN}, the state-of-the-art GAN-based model.}\label{tbl:gau}
\begin{tabular}{cc|ccccccccccccc}
\specialrule{1pt}{1pt}{1pt}
&\multirow{2}{*}{Method} & & \multicolumn{3}{c}{\texttt{Grid}}  & & \multicolumn{3}{c}{\texttt{Gridr}} & & \multicolumn{3}{c}{\texttt{Ring}}  &\\ \cline{4-6} \cline{8-10} \cline{12-14}
& & & $\Pr(\mathcal{F} | \mathcal{S})$ & & $\Pr(\mathtt{T}_{\textrm{test}} | \mathcal{S}')$ & & $\Pr(\mathcal{F} | \mathcal{S})$ & & $\Pr(\mathtt{T}_{\textrm{test}} | \mathcal{S}')$ & & $\Pr(\mathcal{F} | \mathcal{S})$ & & $\Pr(\mathtt{T}_{\textrm{test}} | \mathcal{S}')$ &\\ \specialrule{1pt}{1pt}{1pt}
&$\mathtt{T}_{\textrm{train}}$            & & -3.06          & & -3.06     & & -3.06     & & -3.07     & & -1.70     & & -1.70     &\\ \hline
&\texttt{CLBN}                                     & & -3.68          & & -8.62     & & \underline{-3.76}     & & -11.60    & & \underline{-1.75}     & & \textbf{-1.70}    & \\
&\texttt{PrivBN}                                   & & -4.33          & & -21.67    & & -3.98     & & -13.88    & & -1.82     & & \underline{-1.71}     &\\
&\texttt{MedGAN}                                   & & -10.04         & & -62.93    & & -9.45     & & -72.00    & & -2.32     & & -45.16    &\\
&\texttt{VEEGAN}                                   & & -9.81          & & -4.79     & & -12.51    & & -4.94     & & -7.85     & & -2.92     &\\
&\texttt{TableGAN}                                 & & -8.70          & & -4.99     & & -9.64     & & -4.70     & & -6.38     & & -2.66     &\\
&\texttt{TVAE}                                     & & \textbf{-2.86}     & & -11.26    & & \textbf{-3.41}     & & \textbf{-3.20}     & & \textbf{-1.68}     & & -1.79     &\\
&\texttt{TGAN}                                     & & -5.63     & & -3.69     & & -8.11     & & -4.31     & & -3.43     & & -2.19     &\\ \hline
&\texttt{OCT-GAN(fixed)}                           & & -3.48     & & \underline{-3.47}     & & -4.66     & & \underline{-3.96}     & & -2.39     & & -1.97     &\\ 
&\texttt{OCT-GAN(only\_G)}                          & & -3.41     & & -3.49     & & -4.96     & & -4.03     & & -2.49     & & -1.97     &\\ 
&\texttt{OCT-GAN(only\_D)}                          & & -3.71     & & -3.51     & & -4.85     & & -4.0     & & -2.48     & & -1.98     &\\ \hline
&\texttt{OCT-GAN}                                  & & \underline{-3.32}     & & \textbf{-3.46}     & & -4.90     & & -4.09     & & -2.43     & & -1.98     &\\ \specialrule{1pt}{1pt}{1pt}
\end{tabular}
\end{table*}

\begin{table*}[t]
\small
\centering
\setlength{\tabcolsep}{3pt}
\caption{Likelihood estimation with Bayesian networks. Our \texttt{OCT-GAN} outperforms \texttt{TGAN}, the state-of-the-art GAN-based model.}\label{tbl:bn}
\begin{tabular}{cc|ccccccccccccccccc}
\specialrule{1pt}{1pt}{1pt}
&\multirow{2}{*}{Method} & & \multicolumn{3}{c}{\texttt{Asia}}  & & \multicolumn{3}{c}{\texttt{Alarm}} & & \multicolumn{3}{c}{\texttt{Child}} & & \multicolumn{3}{c}{\texttt{Insurance}} &\\ \cline{4-6} \cline{8-10} \cline{12-14} \cline{16-18}
& & & $\Pr(\mathcal{F} | \mathcal{S})$ & & $\Pr(\mathtt{T}_{\textrm{test}} | \mathcal{S}')$ & & $\Pr(\mathcal{F} | \mathcal{S})$ & & $\Pr(\mathtt{T}_{\textrm{test}} | \mathcal{S}')$ & & $\Pr(\mathcal{F} | \mathcal{S})$ & & $\Pr(\mathtt{T}_{\textrm{test}} | \mathcal{S}')$ & & $\Pr(\mathcal{F} | \mathcal{S})$ & & $\Pr(\mathtt{T}_{\textrm{test}} | \mathcal{S}')$        &\\ \specialrule{1pt}{1pt}{1pt}
&$\mathtt{T}_{\textrm{train}}$            & & -2.23     & & -2.24     & & -10.3     & & -10.3     & & -12.0     & & -12.0     & & -12.8     &     & -12.9    &\\ \hline
&\texttt{CLBN}                & & -2.44     & & -2.27     & & -12.4     & & -11.2     & & -12.6     & & \underline{-12.3}     & & -15.2     &     & -13.9    &\\
&\texttt{PrivBN}              & & \underline{-2.28}     & & \textbf{-2.24}     & & -11.9     & & \underline{-10.9}   & & \underline{-12.3}     & & \textbf{-12.2}     & & -14.7   &     & \textbf{-13.6}    &\\
&\texttt{MedGAN}              & & -2.81     & & -2.59     & &-10.9    & & -14.2     & & -14.2     & & -15.4     & & -16.4     &     & -16.4    &\\
&\texttt{VEEGAN}              & & -8.11     & & -4.63     & & -17.7     & & -14.9     & & -17.6     & & -17.8     & & -18.2     &     & -18.1    &\\
&\texttt{TableGAN}            & & -3.64     & & -2.77     & & -12.7     & & -11.5     & & -15.0     & & -13.3     & & -16.0     &     & -14.3    &\\
&\texttt{TVAE}                & & -2.31     & & -2.27     & & -11.2     & & \textbf{-10.7}     & & \underline{-12.3}     & & \underline{-12.3}     & & -14.7     &     & -14.2    &\\
&\texttt{TGAN}                & & -2.56     & & -2.31     & & -14.2     & & -12.6     & & -13.4     & & -12.7     & & -16.5     &     & -14.8    &\\ \hline
&\texttt{OCT-GAN(fixed)}                & & -2.51     & & -2.27     & & \textbf{-10.7}     & & -11.1     & & -12.5     & & \underline{-12.3}     & & \textbf{-14.5} & & -13.8 &\\ 
&\texttt{OCT-GAN(only\_G)}                & & -2.50     & & -2.27     & & -12.1     & & -11.6     & & \textbf{-12.1}     & & \textbf{-12.2}     & & -14.9 && \underline{-13.7}&\\ 
&\texttt{OCT-GAN(only\_D)}                & & -2.35     & & \underline{-2.26}     & & -11.9     & & -11.1     & & -12.5     & & \textbf{-12.2}     & & -14.8 && \textbf{-13.6}&\\ \hline
&\texttt{OCT-GAN}           & & \textbf{-2.25}     & & -2.27     & & \underline{-10.8}    & & \underline{-10.9}     & & -12.6     & & \underline{-12.3}     & & \underline{-14.6}     &     & -13.9    &\\ \specialrule{1pt}{1pt}{1pt}
\end{tabular}
\end{table*}

\subsection{Classification with Real Data}
\subsubsection{Data}
We consider 5 real-world datasets for classification: \texttt{Adult}~\cite{osti_421279}, \texttt{Census}~\cite{osti_421279}, \texttt{Covertype}~\cite{covertype}, \texttt{Credit}~\cite{fraud}, \texttt{Intrusion}~\cite{intrusion}. \texttt{Adult} consists of diverse demographic information in the U.S., extracted from the 1994 Census Survey, where we predict two classes of high ($>$\$50K) and low ($\leq$\$50K) income. \texttt{Census} is similar to \texttt{Adult} but it has different columns. \texttt{Covertype} is to predict forest cover types from cartographic variables only and was collected from the Roosevelt National Forest of northern Colorado. \texttt{Credit} is for credit card fraud detection, collected from European cardholders in September 2013. \texttt{Intrusion} was used in the international Knowledge Discovery and Data Mining Competition and contains many network intrusion detection samples. \texttt{Adult}, \texttt{Census}, and \texttt{Credit} are binary classification datasets while others are for multi-class classification. We use $\mathtt{T}_{\textrm{train}}$ and $\mathtt{T}_{\textrm{test}}$ to denote training/testing data in each dataset. Their statistics are summarized in Table~\ref{tbl:sta}.

\subsubsection{Evaluation Methodology}
All those datasets provide well separated training/testing sets, and we use them for evaluation. We first train various generative models, including our \texttt{OCT-GAN}, with their training sets. With those trained models, we i) generate a fake table $\mathcal{F}$, ii) train Adaboost~\cite{10.5555/1624312.1624417}, DecisionTree~\cite{10.1023/A:1022643204877}, and Multi-layer Perceptron (MLP)~\cite{10.5555/1162264} with the fake table, and iii) test with $\mathtt{T}_{\textrm{test}}$. We note that these base classifiers have many hyperpameters and we choose the best hyperparameter set for each classifier using the cross-validation method. We use F-1 (resp. Macro F-1) for the binary (resp. the multi-class) classification tasks.

\begin{table}[t]
\small
\centering
\setlength{\tabcolsep}{1pt}
\caption{Classification/regression with real data. `N/A' means severe mode collapse.}\label{tbl:cla}
\begin{tabular}{c|ccccccccccccc}
\specialrule{1pt}{1pt}{1pt}
\multirow{2}{*}{Method} & & \texttt{Adult} & & \texttt{Census} & & \texttt{Credit} & & \texttt{Cover.} & & \texttt{Intru.} & & \texttt{News}                & \\
              & & F1   & & F1    & & F1    & & Macro & & Macro & & $R^2$ & \\ \specialrule{1pt}{1pt}{1pt}
$\mathtt{T}_{\textrm{train}}$            & & 0.669 & & 0.494 & & 0.720 & & 0.652 & & 0.862 & & 0.14                & \\ \hline
\texttt{CLBN}                & & 0.334 & & 0.310 & & 0.409 & & 0.319 & & 0.384 & & -6.28               & \\
\texttt{PrivBN}              & & 0.414 & & 0.212 & & 0.185 & & 0.270 & & 0.384 & & -4.49               & \\
\texttt{MedGAN}              & & 0.375 & & N/A & & N/A & & 0.093 & & 0.299 & & -8.80               & \\
\texttt{VEEGAN}              & & 0.235 & & 0.094 & & N/A & & 0.082 & & 0.261 & & -6.5e6              & \\
\texttt{TableGAN}            & & 0.492 & & 0.358 & & 0.182 & & N/A & & N/A & & -3.09               & \\
\texttt{TVAE}                & & 0.626 & & 0.377 & & 0.098 & & \underline{0.433} & & 0.511 & & -0.20               & \\
\texttt{TGAN}                & & 0.601 & & 0.391 & & 0.672 & & 0.324 & & 0.528 & & -0.43               & \\ \hline
\texttt{OCT-GAN(fixed)}                & & \underline{0.632}     & & 0.370     & & 0.620     & & 0.405     & & 0.453     & & \underline{0.06}      \\ 
\texttt{OCT-GAN(only\_G)}                & & 0.591     & & 0.247     & & 0.660     & & 0.358     & &  \textbf{0.552}     && -4.35    \\ 
\texttt{OCT-GAN(only\_D)}                & & 0.631     & & \textbf{0.436}     & & \underline{0.689}     & & 0.364     & & 0.454     & & -0.17    \\ \hline
\texttt{OCT-GAN}           & & \textbf{0.635} & & \underline{0.402}     & & \textbf{0.695} & & \textbf{0.438}     & & \underline{0.532}     & & \textbf{0.08}                & \\ \specialrule{1pt}{1pt}{1pt}
\end{tabular}
\end{table}

\subsubsection{Experimental Results}
The classification results are summarized in Table~\ref{tbl:cla}. \texttt{CLBN} and \texttt{PrivBN} do not show any reasonable performance in these experiments even though their likelihood estimation experiments with simulated data are not bad. All their (Macro) F-1 scores fall into the category of worst-case performance, which proves potential intrinsic differences between likelihood estimation and classification --- data synthesis with good likelihood estimation does not necessarily mean good classification.

\texttt{TVAE} shows reasonable scores in many cases. In \texttt{Credit}, however, its score is unreasonably low. This also corroborates the intrinsic difference between likelihood estimation and classification.

Many GAN models except \texttt{TGAN} and \texttt{OCT-GAN} show low scores in many cases, e.g., an F-1 score of 0.094 by \texttt{VEEGAN} in \texttt{Census}. Due to severe mode collapse in $\mathcal{F}$, we could not properly train classifiers in some cases and their F-1 scores are marked with `N/A'. However, our proposed \texttt{OCT-GAN}s, including its variations, significantly outperform all other methods in all datasets.

\subsection{Regression with Real Data}
\subsubsection{Data}
We use \texttt{News}~\cite{news} in this experiment, which contains many features extracted from online news articles to predict the number of shares in social networks, e.g., tweets, retweets, and so forth. Therefore, this dataset is good to show the usefulness of our method in web-based applications.

\subsubsection{Evaluation Methodology}
We follow steps similar to the steps in the classification experiment. We use Linear Regression and MLP as base regression models and use $R^2$ as an evaluation metric~\cite{10.5555/1162264}. 

\subsubsection{Experimental Results}
As shown in Table~\ref{tbl:cla}, all methods except \texttt{OCT-GAN} show unreasonable accuracy. The original model, trained with $\mathtt{T}_{\textrm{train}}$, shows an $R^2$ score of 0.14 and our \texttt{OCT-GAN} shows a score close to it. Only \texttt{OCT-GAN} and the original model, marked with $\mathtt{T}_{\textrm{train}}$, show positive scores.

\begin{table}[t]
\centering
\footnotesize
\setlength{\tabcolsep}{1pt}
\caption{Clustering with real data (Silhouette score)}\label{tbl:clu}
\begin{tabular}{c|ccccccccccccccccccccc}
\specialrule{1pt}{1pt}{1pt}
\multirow{2}{*}{Method}   &  & \multicolumn{3}{c}{\texttt{Adult}} &  & \multicolumn{3}{c}{\texttt{Census}} &  & \multicolumn{3}{c}{\texttt{Credit}} &  & \multicolumn{3}{c}{\texttt{Cover.}} &  & \multicolumn{3}{c}{\texttt{Intru.}} &  \\ \cline{3-5} \cline{7-9} \cline{11-13} \cline{15-17} \cline{19-21}
       &  & $\mathtt{T}_{\textrm{train}}$      &     & $\mathtt{T}_{\textrm{test}}$      &  & $\mathtt{T}_{\textrm{train}}$       &     & $\mathtt{T}_{\textrm{test}}$      &  & $\mathtt{T}_{\textrm{train}}$       &     & $\mathtt{T}_{\textrm{test}}$      &  & $\mathtt{T}_{\textrm{train}}$       &     & $\mathtt{T}_{\textrm{test}}$      &  & $\mathtt{T}_{\textrm{train}}$       &     & $\mathtt{T}_{\textrm{test}}$      &  \\ \specialrule{1pt}{1pt}{1pt}
$\mathtt{T}_{\textrm{train}}$            &  & 0.61      &     &0.61      &  & 0.41       &     & 0.39      &  & 0.40       &     & 0.40      &  & 0.16       &     & 0.15      &  & 0.87       &     & 0.86      &  \\ \hline
\texttt{TGAN}            &  & 0.38    &     & 0.60    &  & 0.22     &     & 0.32    &  & 0.40     &     & 0.36    &  & 0.12     &     & 0.11    &  & 0.50     &     & 0.85    &  \\ \hline
\texttt{OCT-GAN(fixed)}   &  & 0.54    &     &   0.60    &  & \textbf{0.53}     &     & 0.41      &  & 0.37     &     & 0.53      &  & 0.13     &     & 0.11      &  & \textbf{0.59}     &     & 0.71      &  \\
\texttt{OCT-GAN(only\_G)} &  & 0.13    &     & 0.53      &  & 0.29     &     & 0.32      &  & \textbf{0.42}     &     & 0.14      &  & \textbf{0.16}     &     & 0.09      &  & 0.34     &     & 0.85      &  \\
\texttt{OCT-GAN(only\_D)} &  & 0.57    &     & 0.62      &  & 0.25     &     & 0.32      &  & 0.36     &     & \textbf{0.54}      &  & 0.08     &     & \textbf{0.15}      &  & 0.27     &     & 0.67      &  \\ \hline
\texttt{OCT-GAN}          &  & \textbf{0.62}    &     & \textbf{0.62}      &  & 0.49     &     & \textbf{0.41}      &  & 0.34     &     & \textbf{0.54}      &  & 0.12     &     & 0.10      &  & 0.45     &     & \textbf{0.86}      &  \\ \specialrule{1pt}{1pt}{1pt}
\end{tabular}
\end{table}


\subsection{Clustering with Real Data}
\subsubsection{Data \& Evaluation Methodology}
We use the 5 classification datasets. With $K=\{|\mathcal{C}|,2|\mathcal{C}|,3|\mathcal{C}|\}$, where $\mathcal{C}$ is a set of class labels, we run $K$-Means++~\cite{10.5555/1283383.1283494} for $\mathcal{F}$. We choose a value of $K$ resulting in the highest Silhouette score~\cite{ROUSSEEUW198753} to find the best $K$. With the found centroids of $\mathcal{F}$, we calculate a Silhouette score of applying the centroids to $\mathtt{T}_{\textrm{train}}$ and $\mathtt{T}_{\textrm{test}}$. A low score for $\mathtt{T}_{\textrm{train}}$ means potential mode collapse or incomplete synthesis.

\subsubsection{Experimental Results}
Table~\ref{tbl:clu} summarizes the results by \texttt{TGAN} and \texttt{OCT-GAN}, the top-2 models for classification and regression, for space reasons. \texttt{OCT-GAN}  outperforms \texttt{TGAN} in almost all cases.

\subsection{Ablation Study}\label{sec:abl}
\subsubsection{Ablation Study Models}
To show the efficacy of key design points in our proposed model, we compare the full model with the following ablation study models:
\begin{enumerate}
    \item In \texttt{OCT-GAN(fixed)}, we do not train $t_i$ but set it to $t_i = \frac{i}{m}$, $0 \leq i \leq m$, i.e., evenly dividing the range $[0,1]$ into $t_0 = 0, t_1 = \frac{1}{m}, \cdots, t_m = 1$.
    \item In \texttt{OCT-GAN(only\_G)}, we add an ODE layer only to the generator and the discriminator does not have it. We set $\texttt{D}(x) = \mathtt{FC5}(\mathtt{leaky}(\mathtt{FC4}(\mathtt{leaky}(\mathtt{FC3}(\bm{h}(0))))))$ in Eq.~\eqref{eq:d:cla}.
    \item In \texttt{OCT-GAN(only\_D)}, we add an ODE layer only to the discriminator and feed $\bm{z} \oplus \bm{c}$ directly into the generator.
\end{enumerate}

\begin{figure}
    \centering
    \subfigure[Education Level]{\includegraphics[trim=0 0 100 45, clip,width=0.49\columnwidth]{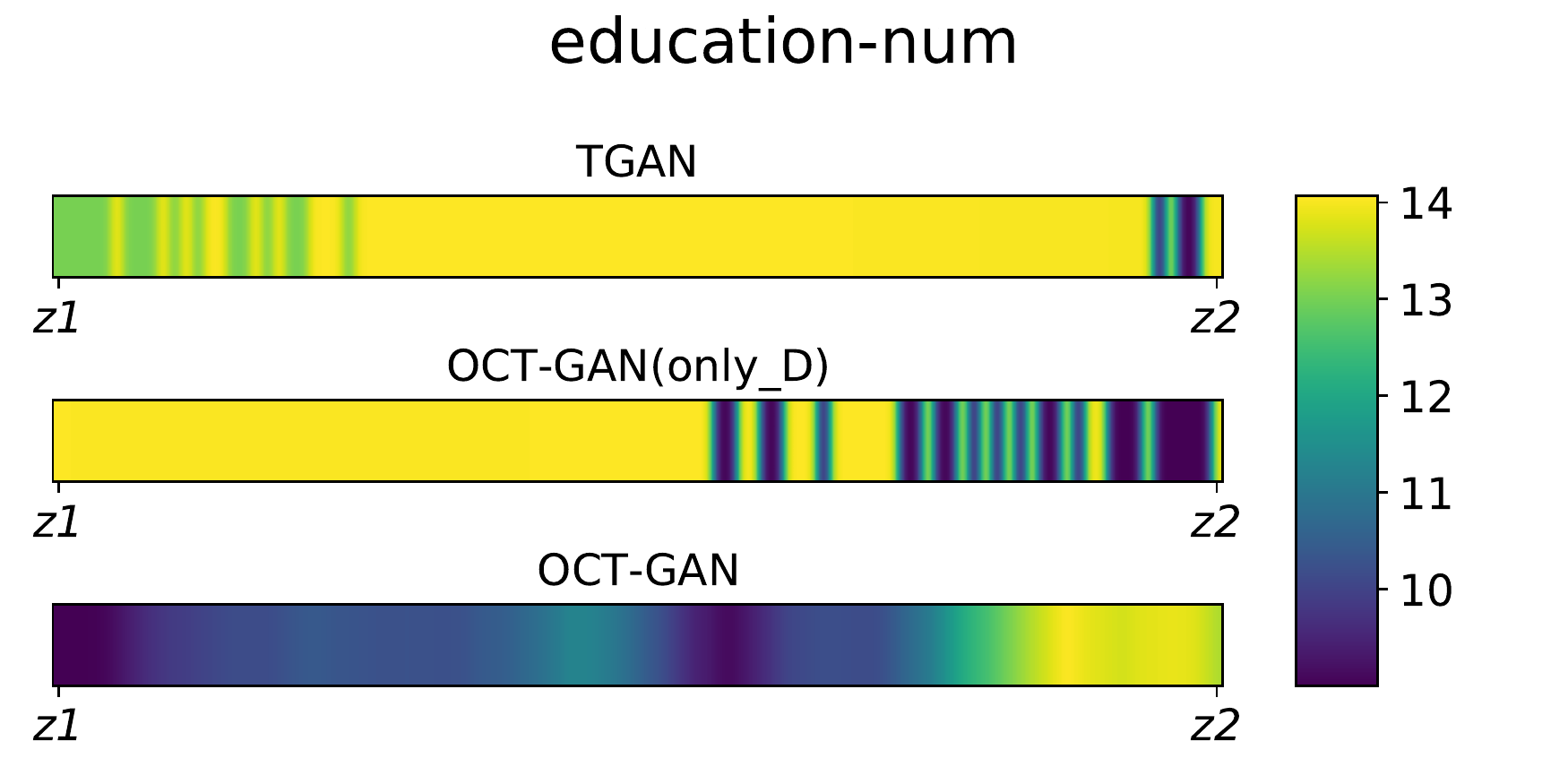}}
    \subfigure[Age]{\includegraphics[trim=0 0 100 45, clip,width=0.49\columnwidth]{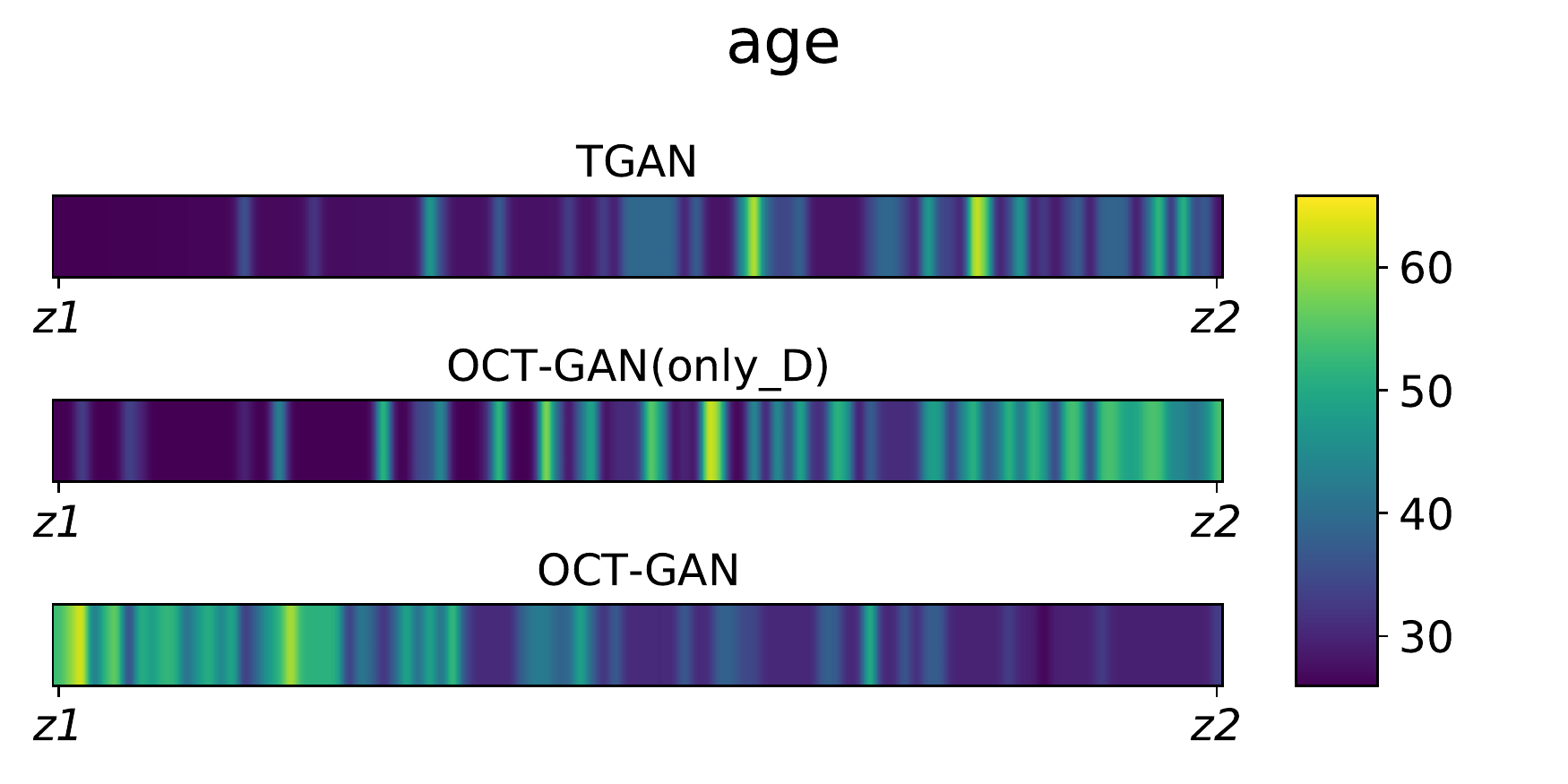}}
    \subfigure[Marital Status]{\includegraphics[trim=0 0 100 45, clip,width=0.49\columnwidth]{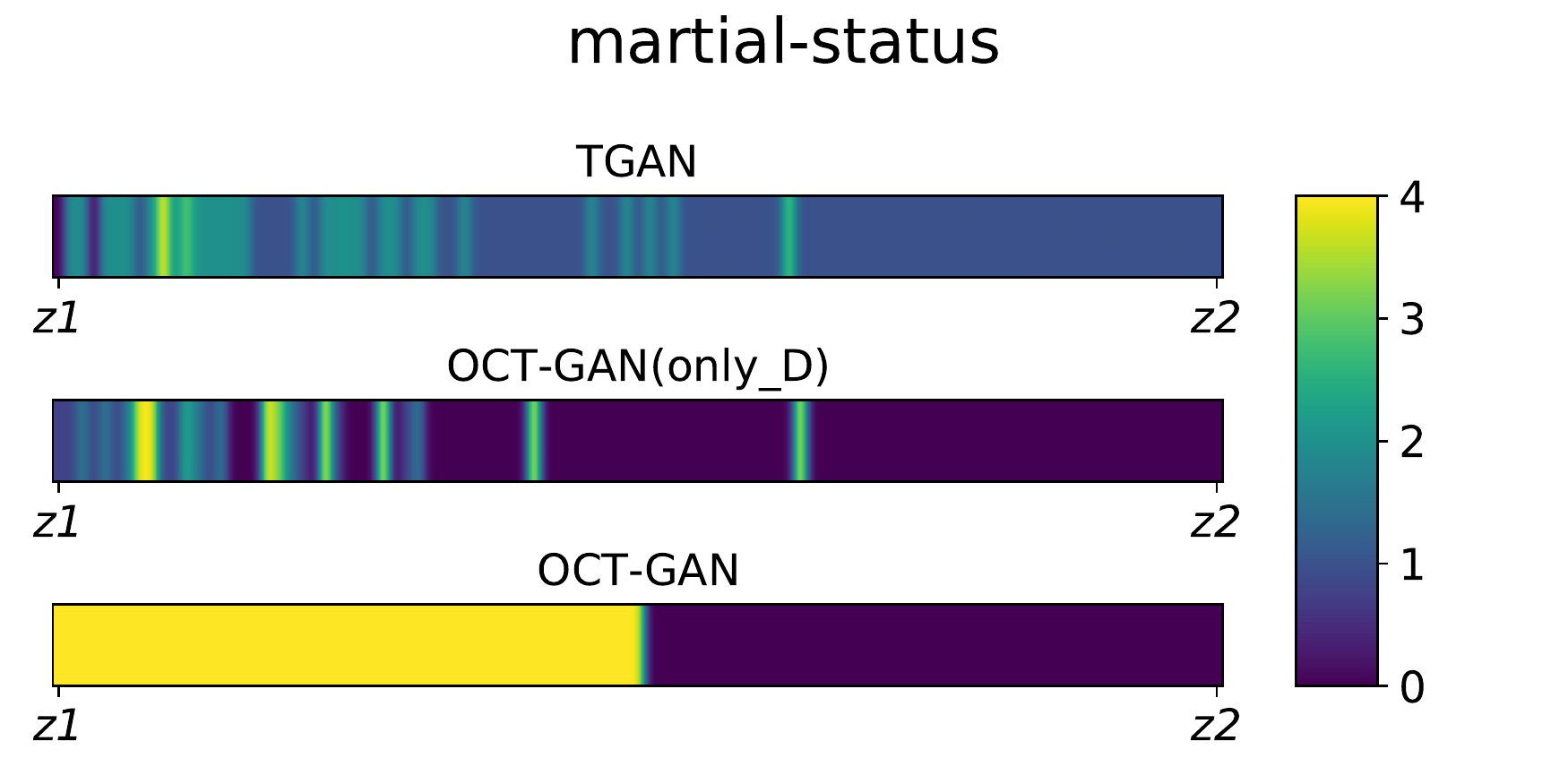}}
    \subfigure[Capital Gain]{\includegraphics[trim=0 0 100 45, clip,width=0.49\columnwidth]{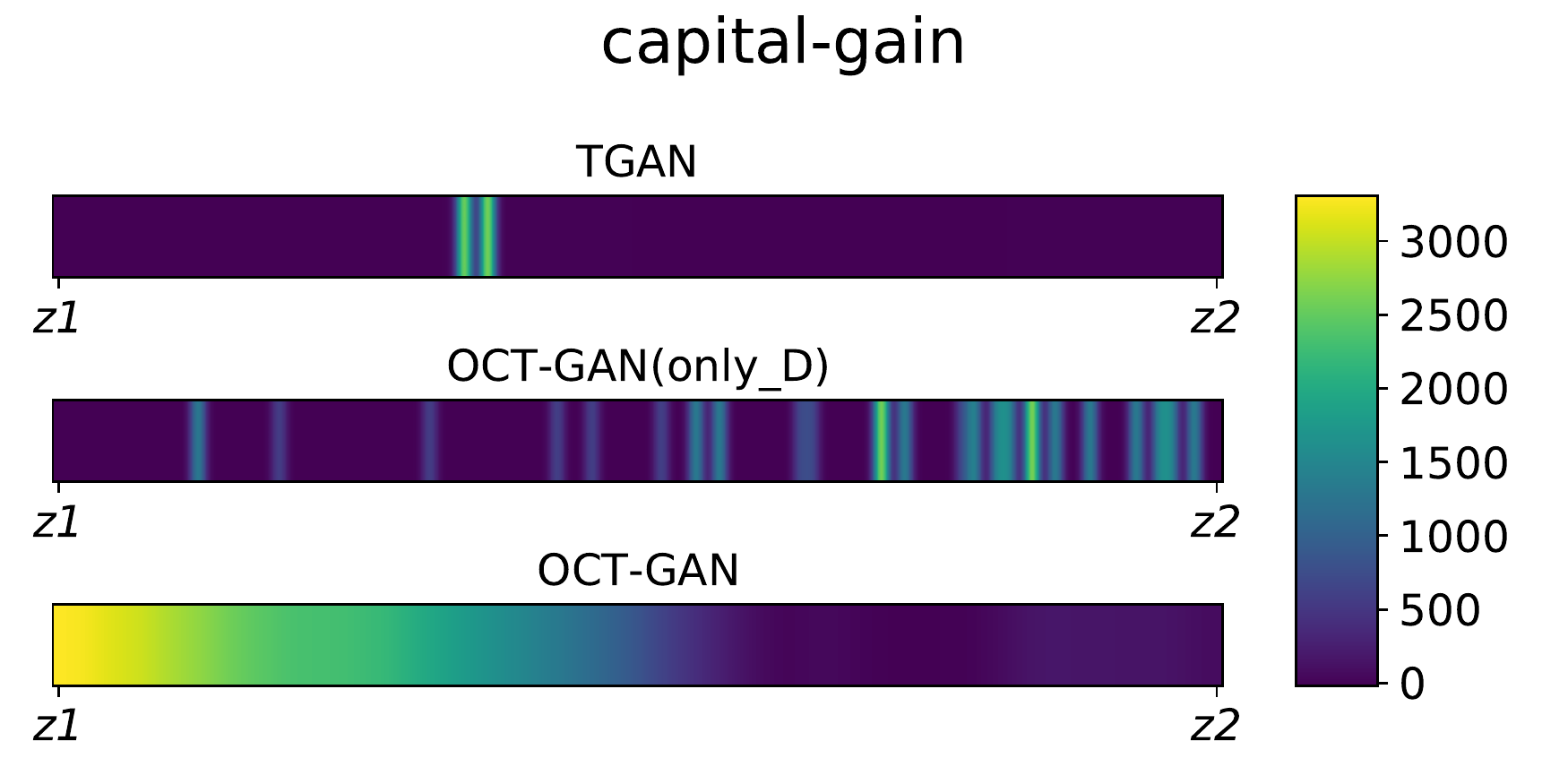}}
    \subfigure[Capital Loss]{\includegraphics[trim=0 0 100 45, clip,width=0.49\columnwidth]{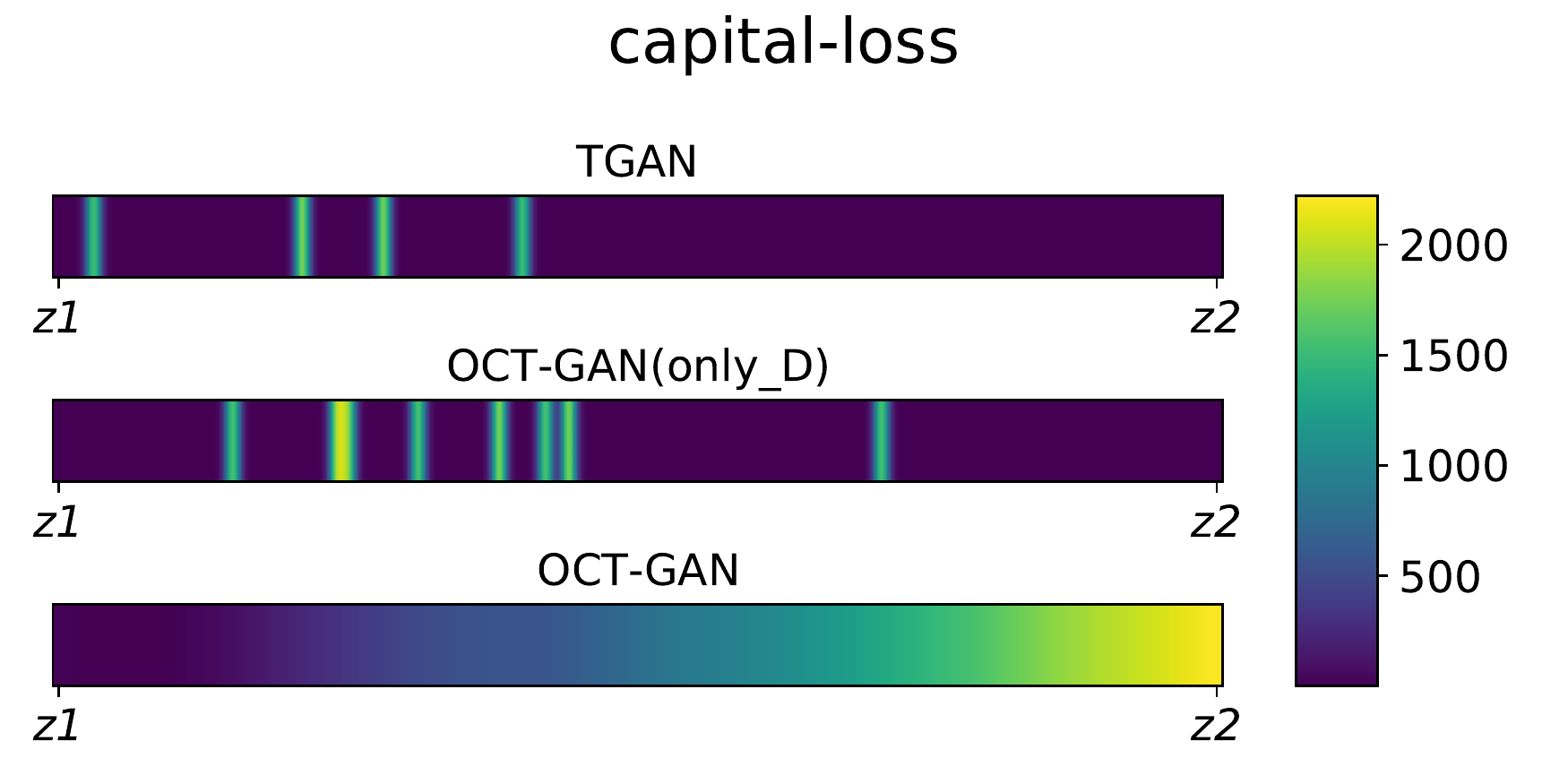}}
    \subfigure[Hours-per-week]{\includegraphics[trim=0 0 100 45, clip,width=0.49\columnwidth]{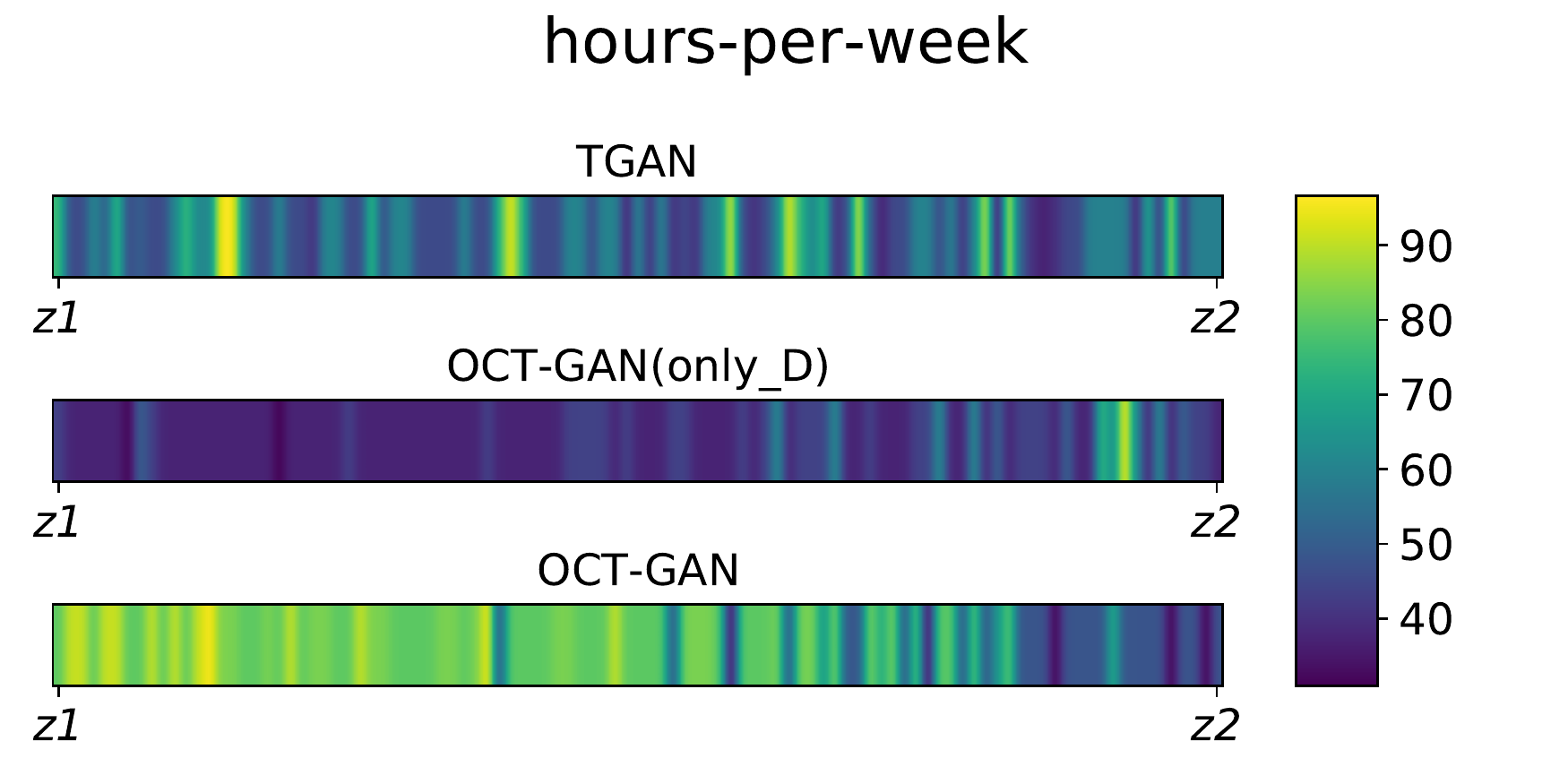}}
    \caption{Interpolation results for important columns in \texttt{Adult} --- purple means the minimum value and yellow means the maximum value in each column. We use the models reported in Table~\ref{tbl:cla} for this visualization.}
    \label{fig:inter}
\end{figure}

\subsubsection{Ablation Study Results}
In Tables~\ref{tbl:gau} to~\ref{tbl:clu}, we also summarize the ablation study models' performance. In Tables~\ref{tbl:gau} and~\ref{tbl:bn}, those ablation study models surprisingly show better likelihood estimations than the full model, \texttt{OCT-GAN}, in several cases. However, we do not observe significant margins between the full model and the ablation study models (even when the ablation study models are better than the full model).

For the classification and regression experiments in Table~\ref{tbl:cla}, however, we can observe non-trivial differences among them in several cases. In \texttt{Adult}, for instance, \texttt{OCT-GAN(only\_G)} shows a much lower score than other models. By this, we can know that in \texttt{Adult}, the ODE layer in the discriminator plays a key role. \texttt{OCT-GAN(fixed)} is almost as good as \texttt{OCT-GAN}, but learning intermediate time points further improves, i.e., 0.632 of \texttt{OCT-GAN(fixed)} vs. 0.635 of \texttt{OCT-GAN}. Therefore, it is crucial to use the full model, \texttt{OCT-GAN}, considering the high data utility in several datasets.

\subsection{Noisy Vector Interpolation}
To further show the efficacy of the ODE-based transformation in the generator, we visualize several interpolation results in \texttt{Adult}. We select two noisy vectors $\bm{z}_1,\bm{z}_2$ and interpolate many intermediate vectors by $e\bm{z}_1 + (1-e)\bm{z}_2$, where $0<e<1$, to generate samples given a fixed random condition vector. In Fig.~\ref{fig:inter}, we show those interpolation results in several columns of \texttt{Adult}. In our observation, \texttt{TGAN} and \texttt{OCT-GAN(only\_D)} show similar interpolation patterns and \texttt{OCT-GAN} can interpolate in a smooth way.

\section{Discussions}
One important discussing point is the difference between the likelihood fitness and the other machine learning experiments. In general, simple models, such as \texttt{PrivBN}, \texttt{TVAE}, and our ablation study models, show better likelihood estimations, and sophisticated models show better machine learning task scores. In real-world environments, however, we think that task-specific data utility is more important than likelihood. Therefore, \texttt{OCT-GAN} can benefit many applications.

However, the data utility of fake tabular data is not satisfactory yet in a couple of cases in our experiments, i.e., \texttt{Covertype} and \texttt{Intrusion} where all methods fail to show a score close to that of the original model marked with $\mathtt{T}_{\textrm{train}}$, which shows the difficulty of data synthesis. They are all multi-class classification datasets. We think there is still a room to improve the quality (utility) of data synthesis for complicated machine learning tasks.

\section{Conclusions}
Tabular data synthesis is an important topic of web-based research. However, it is hard to synthesize tabular data due to its irregular data distribution and mode collapse. We presented a NODE-based conditional GAN, called \texttt{OCT-GAN}, carefully designed to address all those problems. Our method shows the best performance in many cases of the classification, regression, and clustering experiments. However, there is a room to improve for multi-class classification.

\begin{acks}
Jayoung Kim and Jinsung Jeon contributed equally to this research. Noseong Park is the corresponding author. This work was supported by the Institute of Information \& Communications Technology Planning \& Evaluation (IITP) grant funded by the Korea government (MSIT) (No. 2020-0-01361, Artificial Intelligence Graduate School Program (Yonsei University)).
\end{acks}

\bibliographystyle{ACM-Reference-Format}
\bibliography{ref}

\end{document}